%% file: main.tex
\documentclass[10pt,twocolumn,letterpaper]{article}

\usepackage{cvpr}      %
\usepackage{amssymb}
\usepackage{amsmath}
\usepackage{graphicx}
\usepackage{subcaption}

\input{preamble}

\definecolor{cvprblue}{rgb}{0.21,0.49,0.74}
\usepackage[pagebackref,breaklinks,colorlinks,citecolor=cvprblue]{hyperref}

\def\paperID{7802} %
\def\confName{CVPR}
\def\confYear{2024}

\title{CIT: Rethinking Class-incremental Semantic Segmentation %
with a 
Class Independent Transformation}

\author{\fontsize{9.5}{4}\selectfont Jinchao Ge$^{1}$, Bowen Zhang$^{1}$, Akide Liu$^{2}$, Minh Hieu Phan$^{1}$, Qi Chen$^{1}$, Yangyang Shu$^{1}$, Yang Zhao$^{3}$\thanks{Corresponding author: \url{y.zhao2@latrobe.edu.au}} \\
\fontsize{9.5}{4}\selectfont
    $^1$Australian Institute for Machine Learning, The University of Adelaide, $^2$Monash University,
    $^3$La Trobe University \\
}

\begin{document}

\maketitle

\input{sections/0_abstract}

\input{sections/1_intro}

\input{sections/2_relatedwork}
\input{sections/3_method}

\input{sections/4_experiment}

\input{sections/5_conclusion}

\newpage

\input{main.bbl}
\end{document}

%% file: preamble.tex
\usepackage[dvipsnames]{xcolor}

\usepackage{booktabs} %
\usepackage{multirow} %
\usepackage{graphicx} %
\usepackage{arydshln}
\usepackage{colortbl}
\definecolor{mygray}{gray}{.92}

%% file: sections/0_abstract.tex
\begin{abstract}
Class-incremental semantic segmentation (CSS) requires that a model learn to segment new classes without forgetting how to segment previous ones: this is typically achieved by distilling the current knowledge and incorporating the latest data. However, bypassing iterative distillation by directly transferring outputs of initial classes to the current learning task is not supported in existing class-specific CSS methods. Via Softmax, they enforce dependency between classes and adjust the output distribution at each learning step, resulting in a large probability distribution gap between initial and current tasks. We introduce a simple, yet effective Class Independent Transformation (CIT) that converts the outputs of existing semantic segmentation models into class-independent forms with negligible cost or performance loss. By utilizing class-independent predictions facilitated by CIT, we establish an accumulative distillation framework, ensuring equitable incorporation of all class information. We conduct extensive experiments on various segmentation architectures, including DeepLabV3, Mask2Former, and SegViTv2. Results from these experiments show minimal task forgetting across different datasets, with less than 5\% for ADE20K in the most challenging 11 task configurations and less than 1\% across all configurations for the PASCAL VOC 2012 dataset.

         \vspace{-1em} %
\end{abstract}

%% file: sections/1_intro.tex
\vspace{-0.5em}
\section{Introduction}\

\begin{figure}
    \centering
    \includegraphics[width=0.95\linewidth]{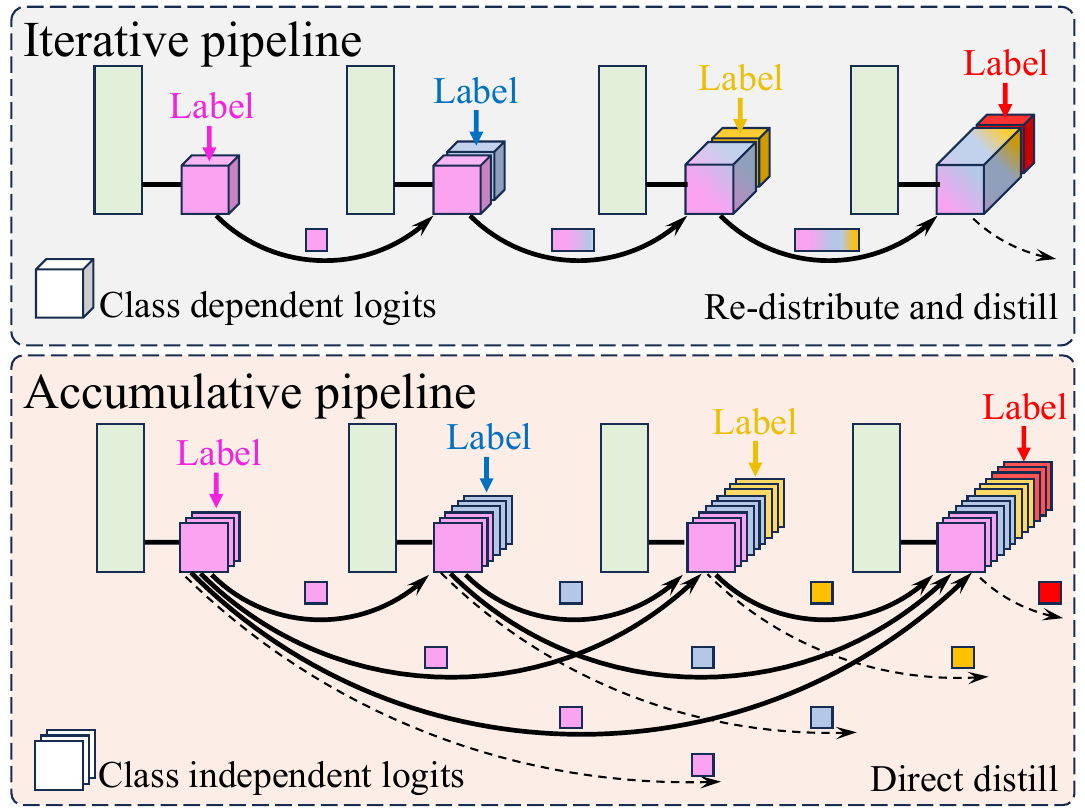}
    \caption{
The top panel illustrates the current iterative distillation pipeline, where knowledge is sequentially transferred and distilled across tasks. This sequential approach can cause an accumulation of errors over time. In contrast,  the lower panel illustrates the use of independent logits, enabling the implementation of our proposed accumulative pipeline. Direct distillation from the source model for each class aids in preserving the integrity of acquired information during the learning sequence. Each square represents the addition of a new task at each stage.
    }
    \label{fig:illustrate}
         \vspace{-1em} %

\end{figure}

Semantic segmentation~\cite{FCN,chen2017rethinking,dosovitskiy2020image,liu2021swin,zheng2021rethinking, chen2018encoder,fu2019dual} is a fundamental task in scene understanding, which aims to assign a label to each pixel. While current deep-learning models~\cite{maskformer,cheng2022masked,zhang2022segvit} perform well on segmentation tasks, they are prone to the problem of catastrophic forgetting~\cite{french1999catastrophic}, which causes the model to lose accuracy on the previous categories as it learns new classes. Therefore, class-incremental semantic segmentation (CSS) \cite{cermelli2020modeling,douillard2021plop,shang2023incrementer, reminder, michieli2019incremental} is an emerging field in computer vision, which aims to optimize the segmentation accuracy in continuously evolving environment.  %
Unlike traditional semantic segmentation models \cite{chen2017rethinking, zhang2023segvitv2,zhang2022segvit, cheng2022masked} trained on a static set of classes, CSS models continually learn to segment new classes without forgetting the previously seen categories. %
Such approaches are critical in real-world applications where new categories are frequently introduced, such as advancements in autonomous vehicles~\cite{Alvarez2012RoadSS, Janai2020ComputerVF} and robotics~\cite{bruce2000fast}.%

CSS methodologies often incorporate techniques like replay strategies \cite{maracani2021recall},
knowledge distillation \cite{douillard2021plop,michieli2021continual, reminder}, and network architecture adaptations~\cite{xiao2014error,serra2018overcoming,mallya2018packnet,li2019learn} that help in retaining knowledge of old classes while efficiently learning new ones. Among these techniques, knowledge distillation (KD) is prominent because it does not rely on storing previously seen data~\cite{reminder}, thus protecting data privacy~\cite{douillard2021plop}. Specifically, when learning new classes, KD methods force the models to produce similar outputs to the previous checkpoint by distilling the probability output between the previous and the current step. %

Despite their remarkable success, existing CSS methods~\cite{douillard2021plop, reminder, shang2023incrementer} suffer from several drawbacks. \textit{Firstly}, these methods adopt conventional semantic segmentation techniques and rely on per-pixel label assignment for segmentation. %
At the output layer, segmentation models often adopt a Softmax operation to generate \textit{class-interdependent} normalized likelihoods, ensuring that the total probability distribution of all learned categories sums to $1$. This design enforces a dependency within a set of target classes. Yet, classes of interest continually change at each CSS step, yielding the mismatch between the probability distribution at different learning steps. This probability mismatch makes the distillation between outputs of different steps non-optimal, potentially impairing the performance of the current model on prior tasks.
\textit{Secondly}, current methods adopt an iterative pipeline, where the predictions of the most recent model are used as the target to re-train the current model on previous categories, as shown in Fig.~\ref{fig:illustrate}.
However, this method causes the model's performance on earlier tasks to deteriorate gradually with each iteration. This degradation, which results from re-distilling the previous tasks repeatedly, leads to incremental forgetting. Moreover, as the model goes through more incremental learning tasks, the accumulated errors and uncertainties on old classes from the learning process %
become more significant. %

To address these limitations in current class-incremental semantic segmentation (CSS) methods~\cite{maracani2021recall,douillard2021plop,michieli2021continual}, we propose a novel approach, called class-independent transformation (CIT) that formulates the segmentation task as a class-agnostic prediction problem. This formulation eliminates the class interdependency enforced by Softmax, which leads to catastrophic forgetting and poor generalization in CSS. %
Our framework introduces a structural transformation, in which predictions from existing semantic segmentation methods are converted into class-independent binary likelihood forms. This conversion is achieved by Sigmoid, acts as a determination function, 
which ensures no compromise in the performance. %
By employing CIT, our framework simplifies the CSS training process, avoiding the class-interdependent normalization constraints. 
For previous categories, we apply a simple distillation loss, utilizing Kullback-Leibler (KL) divergence for the class and mask predictions. For the new categories with ground truth, we adopt the same supervised loss as in the original models, but in the CIT form. Note that, for the distillation process, the target is derived exclusively from supervised training, mitigating the error accumulation seen in previous methods.

In contrast to previous distillation-based CSS, our framework introduces an advanced accumulative knowledge distillation approach, different from the iterative knowledge transfer employed in previous works \cite{reminder,shang2023incrementer,douillard2021plop}. This novel strategy focuses on combining logits of the \textit{source step} checkpoints that correspond to the initial learning stage of each target class. By learning directly from the source model that first learned these target classes, our method avoids the problem of compounded forgetting that often occurs in conventional distillation processes. This accumulative distillation ensures that the current model retains and reinforces knowledge from the initial learning stage, which improves the stability of the base class recognition over time.

Our contributions are as follows:
\begin{itemize}
    \item We empirically show that the interactive distillation and class interdependency enforced by Softmax cause incremental forgetting in continual learning.
    \item We address the forgetting problem by introducing a simple yet effective class-independent transformation (CIT), which reformulates segmentation as a class-agnostic prediction problem. This formulation facilitates a more straightforward accumulative CSS training pipeline that can be integrated into existing semantic segmentation methods.
    \item We demonstrate that the CIT, combined with the accumulative training pipeline, notably mitigates the forgetting issue, resulting in substantially reduced forgetting on the demanding ADE20K datasets compared to conventional CSS methods, and almost negligible forgetting (up to 1\%) across all configurations on Pascal-VOC 2012.

\end{itemize}

%% file: sections/2_relatedwork.tex
\section{Related Work}
\label{sec:related_works}

\noindent\textbf{Continual learning} aims to incrementally learn new tasks without forgetting old ones, which is important for real-world applications where the data and target classes constantly evolve. %
Current methods can be divided into three categories: regularization-based~\cite{dhar2019learning,li2017learning,chaudhry2018riemannian,joseph2022energy} , dynamic architecture-based~\cite{xiao2014error,serra2018overcoming,mallya2018packnet,li2019learn}, and replay-based methods~\cite{Deng2019ClusterAW,shin2017continual,hou2019learning,fernando2017pathnet,rebuffi2017icarl,tiwari2022gcr,zhao2021memory}.
Regularization-based methods~\cite{dhar2019learning,li2017learning} penalize the learning objectives to prevent shifting important parameters. %
Architectural strategies~\cite{xiao2014error,serra2018overcoming,mallya2018packnet,li2019learn,hung2019compacting} dynamically expand new branches or create subnetworks dedicated to new tasks. %
Replay-based methods store data from past tasks and re-train the networks on both new and old data to retain past knowledge. 
They either keep some original data from previous tasks~\cite{Deng2019ClusterAW,hou2019learning} or generate new training samples~\cite{shin2017continual} from old knowledge, using generative models to capture the distribution of past data.

\noindent\textbf{Class-incremental semantic segmentation} (CSS) is an emerging field, which aims to train a continual segmentation model without sacrificing the segmentation accuracy on old tasks. %
Various methods have been proposed to address these problems, which can be grouped into two categories: replay-based methods~\cite{maracani2021recall}, and
distillation-based methods~\cite{douillard2021plop,michieli2021continual}.
Replay-based RECALL~\cite{maracani2021recall} uses generative adversarial networks and web-crawled data to regenerate learning samples from old classes when training the new tasks. 
Distillation-based methods~\cite{cermelli2020modeling,douillard2021plop,reminder,shang2023incrementer} are more prominent in CSS literature since they bypass the need to store old data. These methods transfer the knowledge from the previous checkpoint to the current model, preserving the accuracy of the past tasks.
PLOP~\cite{douillard2021plop} proposes a multiscale distillation, which extracts features at different scales, capturing both long-range and short-range relationships.
MiB~\cite{cermelli2020modeling} allows the old model to predict background pixels as one of a new class in the current task, aiming to resolve the background shift problem.
Prototype-based methods, such as SDR~\cite{michieli2021continual}, leverage prototype matching and contrastive learning to enhance feature robustness in continual semantic segmentation (CSS). REMINDER ~\cite{reminder} features a novel Class Similarity Knowledge Distillation (CSW-KD) method to selectively reinforce the learning of old classes similar to new ones, thereby improving class retention and recognition. Incrementer~\cite{shang2023incrementer}, in contrast, introduces class tokens to the decoder for new class learning and a targeted knowledge distillation strategy to enhance learning plasticity. However, current distillation methods adopt an iterative approach, which only distills the knowledge of the latest checkpoint. While being effective for the most recently seen classes, the accuracy of more former tasks gradually drops at each step, compounding the forgetting issues.

\noindent\textbf{Class agnostic segmentation} methods~\cite{maskformer, strudel2021segmenter, knet,segvitv2,cheng2022masked,zhang2022segvit} have introduced a two-step strategy: image partitioning and class-agnostic region classification. Image partitioning uses learnable tokens as queries and learns to generate object masks. %
Simultaneously, the tokens from region classification are input into a linear classifier to determine the presence of the object in the associated mask. Interestingly, existing literature primarily focuses on adapting class-specific segmentation models, overlooking the effectiveness of class-agnostic segmentation paradigms for continual segmentation, which are now dominating due to their superior performance. 

This paper identifies the cross-task distribution tweaking problem introduced by class-specific segmentation methods, thus taking the first step in exploring class-agnostic prediction for effective CSS.

%% file: sections/3_method.tex
\vspace{-0.5em}
\section{Problem statement}
Class-incremental semantic segmentation (CSS) is a learning paradigm focused on training a segmentation model over multiple tasks $t \in [0, 1, ..., T]$, with the aim of achieving optimal performance across all stages, including previously encountered tasks. %
For each task $t$, we are given a dataset $D^t$, 
comprising labeled data pairs $(X^t, Y^t)$, where $X^t$ and $Y^t$, respectively, denote an image of size $H \times W$ and its ground-truth segmentation map. The ground truth $Y^t$ only consists of the classes of the current task $C^{t}$, while all other classes (i.e., old classes $C^{0:t-1}$ or future classes $C^{t+1:T}$) are assigned to the background class. %

The primary objective of CSS is to develop a training paradigm that achieves performance comparable to a joint model trained on the entire, $(X, Y)$ utilizing all categories in $C$. The discrepancy in performance between the model trained after completing $T$ tasks and the joint model trained with access to all categories is commonly used to quantify continual learning performance. 
         \vspace{-0.5em} %

\section{Methodology}

\begin{figure*}[htbp]
    \centering
    \includegraphics[width=0.68\textwidth]{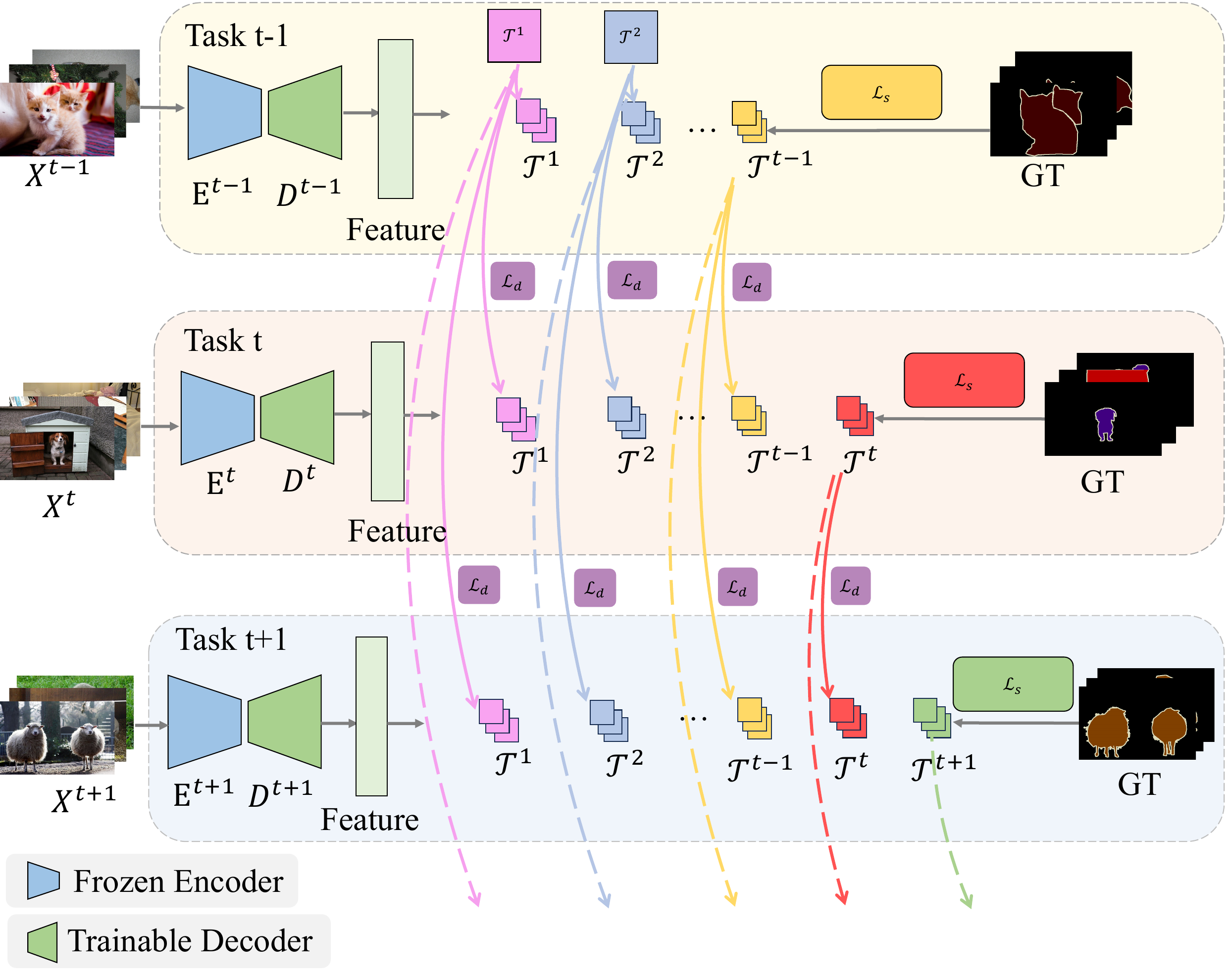}
    \caption{CIT Overview. At each task $t$, it trains its decoder to incorporate the latest class label, enabling it to produce segmentation predictions for the newly introduced classes. Initially, the teacher model $\mathcal{T}^{t-1}$ generates a pseudo-label for the existing classes up to t-1. Subsequently, the model undergoes training utilizing two key components: a feature knowledge distillation loss $\mathcal{L}_{distill}$ ($\mathcal{L}_d$) and a supervised $\mathcal{L}_{supervised}$ ($\mathcal{L}_s$).
    }
    
    \label{fig:cls_agnostic}
             \vspace{-1.5em} %
\end{figure*}

\subsection{Problem analysis}
\label{sec:m_problem_analysis}
Class-incremental semantic segmentation (CSS) is more complex than traditional class incremental learning (CIL) which focuses on classifying a single object in an image.  In traditional CIL, images $X^t$ of the task $t$ only contain a single object from a category belonging to $C^t$. Conversely, CSS concerns images $X^t$ containing multiple objects from categories within the current task $C^t$, and from other tasks as a background class. 

The most common approach to CSS is distillation~\cite{hinton2015distilling}. In an image $X^t_{i}$ depicting a complex scene, there are often objects that belong to previously learned categories $C^{0,t-1}$ %
along with objects that belong to the labeled categories $C^t$. The objects belonging to $C^t$ have labels, whereas, for the objects in $C^{[0, t)}$, the only source of information that can be acquired is the model outputs from previous stages. To enable continuous learning, it is natural to design a framework to `distill' predictions made by the models from previous learning stages into the current model. This approach, known as model distillation, is widely adopted by many existing CSS methods.

Most CSS methods stem from prior semantic segmentation techniques and thus rely on per-pixel classification.  They therefore incorporate a Softmax operation in their final stage to predict a normalized likelihood for each pixel, ensuring that the total probability of all learned categories amounts to $1$.
As the model progresses to task $t$, new categories $C^{t}$ are introduced. Theoretically, the supervision of the logits associated with the newly added categories should be distinct from those of the previously learned categories. However, due to the nature of Softmax, which is tailored to generate a new probability distribution for the expanded category group, the supervision of the newly added categories unavoidably impacts the previously learned groups.

Hence, a simple distillation approach may face challenges in balancing the gradients between distillation from previous stages and the adjustments from the newly added labeled supervision. Additional considerations and strategies are necessary to effectively harmonize and integrate diverse distributions while preserving previously acquired knowledge in CSS methods.

\subsection{Gaps in the CSS training}
\label{sec:m_gap}

In distillation-based methods, forgetting stems from the fact that the output of the previous stage model is the exclusive source of information about previously learned classes. One of the inherent limitations of segmentation models, however, is that they are unable to achieve 100\% training accuracy. As a result, the pseudo-labels generated from the previous model's predictions regarding previously learned categories inevitably contain errors.  This was demonstrated by~\cite{shang2023incrementer} when addressing the `forgetting' issue through a simple pseudo-label supervision approach.

Instead of resorting to pseudo labeling, characterized as `hard' labeling, the method proposed by ~\cite{reminder} involves manipulating the probability distribution of previous models and delicately engineering category probabilities of previous classes $C^{0:t-1}$ %
which requires the use of manually crafted thresholds and rules (e.g., suppressing the probability of similar categories). Regrettably, the manipulation is insufficient and leads to mathematical instability.

Furthermore, the training accuracy not only on ground truth but also in the distillation process, cannot reach perfection. {Yet, previous methods adopt iterative distillation, which distills only the latest model. As such, the accuracy on early steps $C^m$, where $m << t$ significantly drops at every new step after $m$. Iterative distillation compounds the forgetting effects of early tasks. This design stems from the \textit{class inter-dependent} output of current Softmax-based methods. The probability mismatch between different groups prevents current methods from directly distilling the outputs of the source model at early step $m$ to the current model at step $t$.} 
Consequently, these methods simply concatenate new category sets to the previously merged category sets, thereby compounding the errors caused by the distillation process, especially with early tasks being distilled multiple times. This sequential behavior ultimately exacerbates the issue.

\vspace{-0.5em}
\subsection{Class Independent Transformation (CIT)}
\vspace{-0.5em}
\label{subsec:CIT}
According to our analysis, a major obstacle lies in the interdependent prediction representations during both training and inference. To address this challenge, a method that can convert predictions of existing semantic segmentation methods into a class-independent form would be pivotal.
The transformation from class-dependent predictions to class-independent predictions is depicted in Fig. ~\ref{fig:cls_agnostic}.

In fully supervised semantic segmentation methods, two primary prediction forms are prevalent. Beginning with FCN~\cite{FCN}, predictions are typically generated by a Linear layer combined with a Softmax operation, setting a standard for its simple and effective structural design. The dominance of the Softmax prediction form continues until the emergence of DETR~\cite{DETR}, which decouples feature map logits into a class prediction and a binary mask decoupled form. The DETR structure, designed to unify object detection, instance segmentation, and semantic segmentation, exhibits redundancy for the specific task of semantic segmentation.

As shown in Fig.~\ref{fig:cls_agnostic}, we present a comprehensive illustration of our accumulated pipeline. In our detailed exploration, we first adjust the number of queries from a fixed constant to the number of classes. Each token is directly assigned a category, eliminating the need for bipartite matching to assign ground truth. Given that semantic segmentation entails deterministic categories, unlike object detection and instance segmentation, which involve uncertain numbers of objects, this adjustment proves essential. Moreover, with each token assigned to a specific category, class predictions as outputs are no longer necessary; instead, only a binary prediction indicating the presence of a specific class in the image is required. The binary mask prediction remains unchanged. Lastly, to ensure complete independence in predictions, all self-attentions in the transformer decoder layers, which apply self-attention along all the queries, are removed. Empirically, this adjustment has demonstrated no adverse impact on the performance of the semantic segmentation method. Through this transformation, each query becomes entirely independent of other categories.

This method can also be applied to the Softmax prediction forms used in earlier semantic segmentation methods. In order to align with the low computational and parameter cost of the traditional Softmax predictions, we utilize a minimal number of parameters and operations. Consequently, the omission of the FFN module and the Linear operation within the Linear Function module does not significantly impact performance in practice, while introducing only negligible computational and parameter costs.
\vspace{-0.5em}
\begin{equation}
    \mathcal{L}oss_{all}  =\left\{
    \begin{array}{ll}
    \mathcal{L}_{distill}({Pred}^{0,t-1},  {Pred}^t) & \text{for c $\in  C^{0,t-1}$ } \\
    \mathcal{L}_{supervised}({Labels},  {Pred}^t) & \text{for c $\in  C^{t}$ } 
\end{array}
\right. \\
\label{eq:loss}
\end{equation}

Applying the class-independent transformation (CIT) to prior semantic segmentation methods streamlines the process of conducting CSS. As demonstrated in Eq.~\ref{eq:loss}, during training, we gather all predictions from previous stages, denoted as $Pred$, while the prediction specific to the current task is indicated as $Pred^t$, and predictions of previous tasks as $Pred^{0,t-1}$. Within $Pred^t$, both the recently added categories $C^t$ and all previously learned categories $C^{0,t-1}$ are encompassed. 
\vspace{-0.5em}
\begin{equation}
    \mathcal{L}_{distill}  = \lambda_1 \mathcal{KD}({class}) + \lambda_2 \mathcal{KD}({mask})
    \label{eq:distill}
\end{equation}

For the prior categories, the loss consists of a simple distillation loss as demonstrated in Eq.~\ref{eq:distill} (where KD represents KL divergence, and $\lambda_1$ and $\lambda_2$ denote the weighting factors for the respective KD terms
). Concerning the newly introduced categories $C^t$, given the availability of ground truth labels, they adopt the same supervised loss as their original models in the CIT forms.
It is important to note that within $\mathcal{L}_{distill}$, the target is derived from all previous $Pred^{[0,t-1]}$, with each $Pred$ exclusively extracting the logits of categories that have ground truth labeled training. Consequently, all the logits in the distillation process originate from freshly supervised training, thereby mitigating the accumulation of errors that previous methods incur through repeated distillation. With the CIT, previous semantic segmentation methods can be converted into the same class-independent prediction form and apply the loss as in Eq.~\ref{eq:loss} for CSS training.

%% file: sections/4_experiment.tex
\section{Experiment}
\input{tables/exp_ade20k}

\begin{figure*}[ht]
    \centering
    \resizebox{0.95\linewidth}{!}{%
    \begin{subfigure}[b]{0.33\textwidth}
        \includegraphics[width=\textwidth]{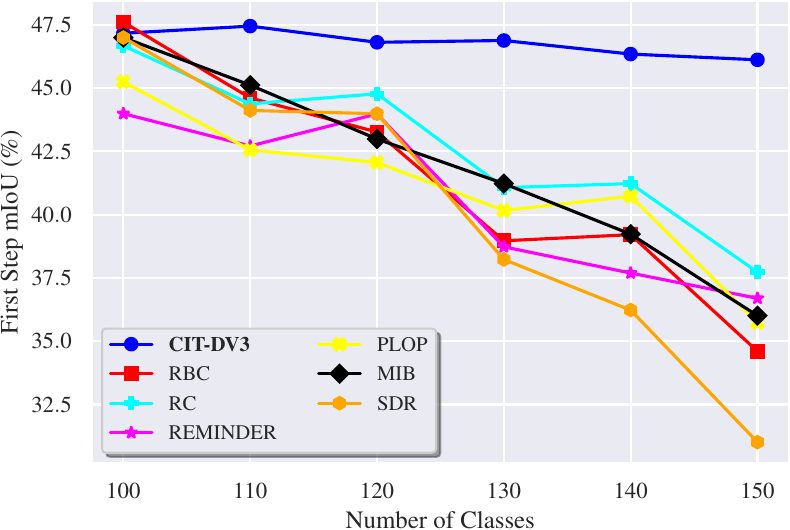}
        \caption{DeepLabV3 100-10 (6 tasks)}
        \label{fig:fig1}
    \end{subfigure}
    \hfill
    \begin{subfigure}[b]{0.33\textwidth}
        \includegraphics[width=\textwidth]{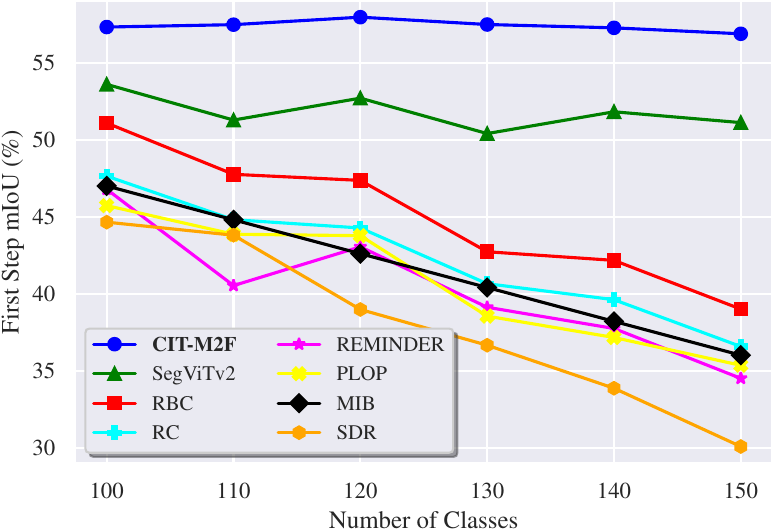}
        \caption{Mask2Former 100-10 (6 tasks)}
        \label{fig:fig2}
    \end{subfigure}
    \hfill
    \begin{subfigure}[b]{0.33\textwidth}
        \includegraphics[width=\textwidth]{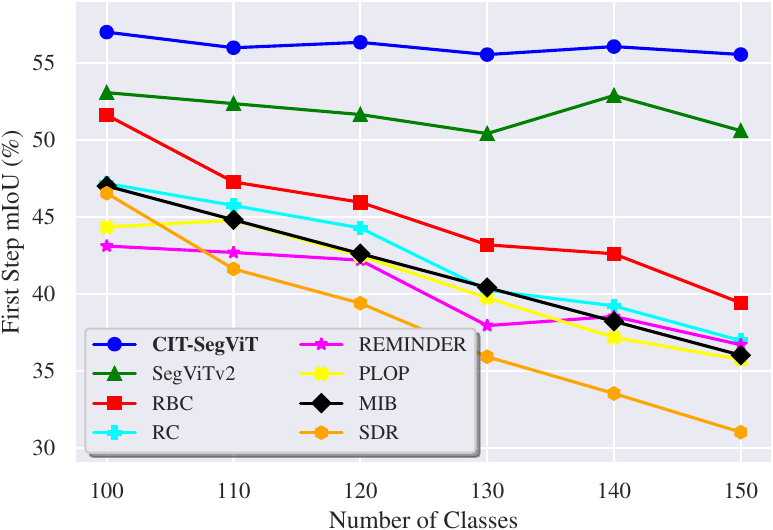}
        \caption{SegViTv2 100-10 (6 tasks)}
        \label{fig:fig3}
    \end{subfigure}
    }

    \resizebox{0.95\linewidth}{!}{%
    \begin{subfigure}[b]{0.33\textwidth}
        \includegraphics[width=\textwidth]{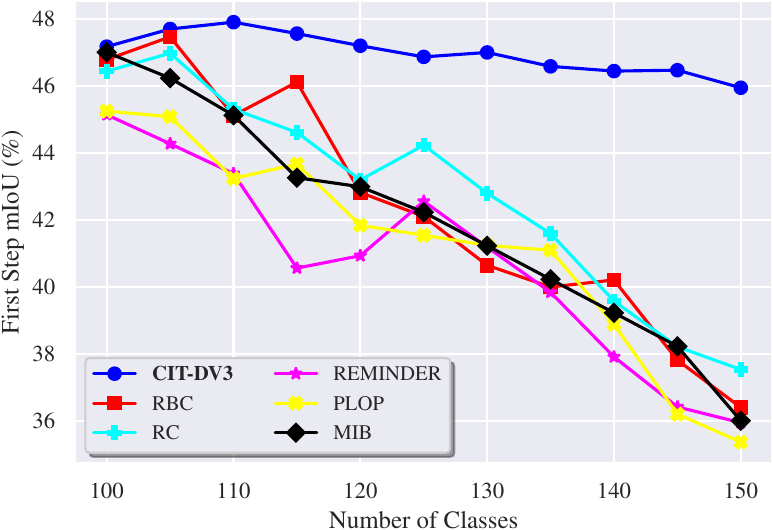}
        \caption{DeepLabV3 100-5 (11 tasks)}
        \label{fig:fig4}
    \end{subfigure}
    \hfill
    \begin{subfigure}[b]{0.33\textwidth}
        \includegraphics[width=\textwidth]{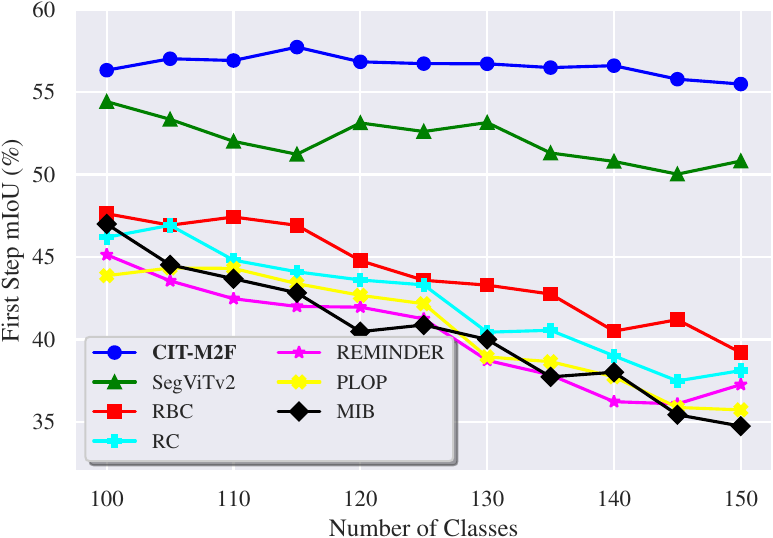}
        \caption{Mask2Former 100-5 (11 tasks)}
        \label{fig:fig5}
    \end{subfigure}
    \hfill
    \begin{subfigure}[b]{0.33\textwidth}
        \includegraphics[width=\textwidth]{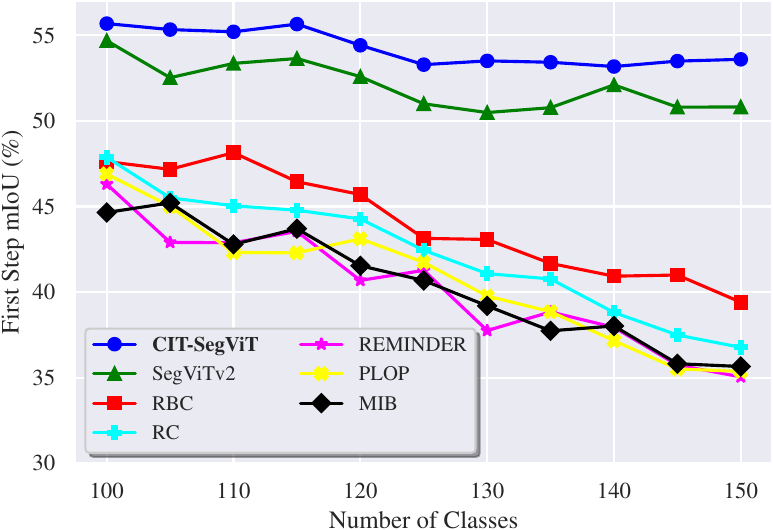}
        \caption{SegViTv2 100-5 (11 tasks)}
        \label{fig:fig6}
    \end{subfigure}
    }

\caption{This figure compares the extent of forgetting by evaluating base tasks in terms of mIoU vs the training step count in two challenging protocols: 100-10 and 100-5. (a) and (d) utilize \textbf{DeepLabV3}, offering a comparative analysis with preceding methodologies. (b) and (e) are based on \textbf{Mask2Former}, juxtaposed against prior techniques.  c) and (f) deploy \textbf{SegViTv2}, compared with earlier methods. 
}
    \label{fig:six_figures_ade}
             \vspace{-1em} %

\end{figure*}

\subsection{Experimental Setup}
\textbf{Datasets.} Our research includes comprehensive experiments on two key datasets: Pascal VOC \cite{everingham2015pascal} and ADE20K \cite{zhou2017scene}. The Pascal VOC dataset comprises 20 distinct foreground classes, featuring a training set of 10,582 images and a testing set of 1,449 images. In contrast, the ADE20K dataset encompasses a broader range of 150 classes, providing 20,210 images for training purposes and 2,000 images for testing. 
These datasets serve as critical benchmarks for evaluating the performance and robustness of our experimental models.

\input{tables/exp_voc12}

\noindent\textbf{Incremental Protocols.}
\label{sec:incremental protocols}

We conduct experiments on the Pascal-VOC 2012 and ADE20K datasets, adhering to the protocols outlined in \cite{cermelli2020modeling, douillard2021plop}. We begin by training the chosen network architecture in the first task to recognize an initial set of classes \(C^{0}\) using the corresponding training data \(D^{0}\). Subsequently, we perform a series of incremental tasks, indexed by \(t \in \{1, 2, \ldots, T \}\), where the model learns a set of classes \(C^{t}\) for each task. For the \textit{Pascal-VOC 2012} dataset, we use the 15-5-2, 19-1-2, and 15-1-6 settings, denoting the ${Num}_{init}$-${Num}_{incr}$-${Num}_{tasks}$. In the \textit{ADE20K} dataset, we explore the settings 100-50-2, 50-50-3, 100-10-6, and 100-5-11.

To assess the incremental learning capability, the dataset is partitioned into various subsets based on class differentiation for multistep learning. CSS \cite{cermelli2020modeling} introduces two distinct division settings: \textit{disjoint} and \textit{overlap}. In the \textit{disjoint} setting, the dataset for each learning phase solely comprises the first step of previously learned classes \(C^{t}\) and the rest of classes \(C^{t+1:T}\), excluding any future classes. Here, the previously learned classes are designated as background. Conversely, in the \textit{overlap} setting, the dataset for each phase additionally includes future classes, offering a representation more aligned with real-world scenarios.

\noindent\textbf{Evaluation metrics.} We utilize the mean Intersection over Union (mIoU). We assess the mIoU for the initial step of classes \(C^{0}\) after retraining steps to evaluate stability, the mIoU for subsequent classes \(C^{t:T}\) to assess plasticity, and the mIoU across all classes to determine overall performance. These experiments aim to explore the incremental learning capabilities and the efficacy of our method under diverse and progressively complex scenarios.

\noindent\textbf{Implementation Details.} 
Our experiments utilize the MMSegmentation toolbox \cite{mmseg2020}, built on PyTorch 2.0.0. In our approach, standard data augmentation techniques, including random flipping, cropping, and scaling (ranging from 0.5 to 2), are applied. The optimization process is conducted using AdamW with a momentum of 0.9. We adopt specific image cropping dimensions for each dataset: \(512 \times 512\) for ADE20k and \(480 \times 480\) for Pascal VOC.
Concerning network architecture, DeepLabV3 \cite{chen2017rethinking}, SegViT \cite{zhang2022segvit, zhang2023segvitv2}, and Mask2Former \cite{cheng2022masked} serve as our foundational models. For the initial class stage training \(C^{0}\), we adhere to the original training schedules. However, to mitigate overfitting, the number of training iterations for subsequent classes \(C^{t+1:T}\) is reduced. 
In the context of incremental learning at task \(t\), the model for the current task (\(t\)) is initialized using the parameter settings from the previous task (\(t-1\)). This strategy is designed to emulate the Class-Specific Segmentation (CSS) setting in realistic scenarios.

\subsection{Comparison with the State-of-the-Art}

\paragraph{Evaluations on the ADE20K Dataset.}
Our extensive experimental analysis on the ADE20K dataset involved a broad spectrum of network architectures, including both CNN and Transformer-based models. The empirical results, as tabulated in Tab.~\ref{tab:CIT}, reveal that our method significantly surpasses previous state-of-the-art (SOTA) methods in all four evaluated protocols. Notably, improvements of up to 21\% and 19\% were observed in DeeplabV3-based and SegViTv2-based frameworks, respectively. Our method's efficacy was particularly evident in short-term continual learning scenarios, such as 100-50 (2 tasks) and 50-50 (3 tasks) protocols. For instance, the DeeplabV3-based model exhibited a negligible decrement in performance, attaining mIoU scores of 43.02\% and 42.54\%, closely rivaling Oracle's 43.40\% mIoU. In the SegViTv2 configuration, the performance drop was less than 0.5\% mIoU, affirming the robustness of our approach.

In the more challenging long-term learning scenarios, such as the 100-10 (6 tasks) protocol, our method continued to demonstrate minimal forgetting, aligning closely with Oracle's performance. In the Mask2former-based setting, our system achieved a remarkable 56.15\% mIoU, nearly matching the joint oracle training standard of 56.32\% mIoU. For the new classes on 100-5 setting (Tab.~\ref{tab:1-2}), our CIT-SegViTv2 achieves 37.78\% mIoU, while SegViTv2~\cite{segvitv2} only obtains 31.26\% mIoU. This shows that transforming the output into class-independent probability enables effective learning of new classes.

\vspace{-1.3em}
\paragraph{Performance on the Pascal VOC Dataset.}
Extending our analysis to the Pascal VOC dataset, we assessed our methodology across three distinct protocols, as outlined in Tab.~\ref{tab:1-2}. In the shorter-term 19-1 and 15–5 settings, our model outperformed the Incrementor by up to 7\% in overall metrics. Our method was particularly effective in minimizing forgetting, evidenced by a retention rate of 86.22\% against the Oracle's 86.78\% in the 19-1 configuration, and 86.61\% and 86.96\% in the 15–5 settings. In longer-term learning, the resilience of our approach was notable, with a performance decline of less than 5\% in the initial phase. For the new classes 16-20 on the challenging 15-1 setting, our CIT-DeepLabV3 achieves a similar performance as the Oracle trained on joint data: 77.32\% versus 78.83\%. Previous Softmax-based methods lead to output probability mismatch between tasks, impairing the model performance during distillation. In contrast, our proposed CIT eliminates the class interdependency, thus improving the training efficiency of CSS on both old and new categories.

\vspace{-1.0em}
\paragraph{Resilience to incremental forgetting.} This section monitors the performance drop of the initial task as the model continually learns new tasks on ADE20k (\cref{fig:six_figures_ade}) and Pascal VOC (\cref{fig:image1_label_voc}, and \cref{fig:image2_label_voc}). The clear drop of previous methods indicates the \textit{incremental forgetting} issue in the iterative distillation pipeline. In contrast, our proposed CIT manifests nearly \textbf{zero} forgetting. This shows that by excluding the class interdependency and enabling accumulative distillation, our method effectively preserves the knowledge of the base classes, yielding state-of-the-art performance across 8 challenging CSS settings.

\subsection{Ablation study}
\subsubsection{Effect of CIT} 
We initially demonstrated that Class-independent Transformation (CIT) can be widely applied across various existing semantic segmentation methods without adversely affecting their performance. The corresponding results can be found in Tab.~\ref{tab:CIT_E}. For instance, DeepLabV3, a widely-used method, adopts a pyramid CNN-based backbone like ResNet and utilizes softmax for prediction. The current state-of-the-art method, Mask2Former, employs a pyramid transformer with a Swin backbone and utilizes a prediction format similar to DETR. Similarly, SegViT, another state-of-the-art method, is specifically designed to utilize a plain vision transformer backbone.

\begin{table}[h!]
\centering
\resizebox{0.95\linewidth}{!}{%
\begin{tabular}{lccc}
\toprule
Method & DeepLabV3 & Mask2former & SegViTv2 \\
\midrule
Orig mIoU    & 42.7 & 53.9 & 53.5 \\
CIT mIoU     & 43.4 (+0.7) & 53.5 (-0.4) & 53.7 (+0.2) \\
\midrule
Param diff & +0.0\% & -2.8\% & -4.5\% \\
GFLOPs diff  & +2.4\% & -0.4\% & -1.9\% \\
\bottomrule
\end{tabular}
}
\caption{
Comparison of the performance of established semantic segmentation methods following integration with the CIT module. The experiment was carried out using the ADE20K dataset.}
\label{tab:CIT_E}
         \vspace{-1em} %

\end{table}

\vspace{-0.2em}
These three methods were chosen as they representatively illustrate the application of our CIT module. Upon application of CIT, it is evident that there are no significant changes in terms of mIoU, number of parameters, and computational cost. Consequently, we confidently assert that the CIT process does not detrimentally impact the overall performance of an existing semantic segmentation method.

\vspace{-1em}
\subsubsection{Pseudo labeling VS Soft labeling}
Pseudo-labeling is a widely used approach for transferring knowledge from previous tasks to the current one. In Tab.~\ref{tab:pseudo}, we maintain the 100-50 configuration, covering two tasks, to mitigate potential performance drops across multiple tasks. It becomes clear that pseudo-labeling, which inherently leads to information loss, shows noticeable performance degradation compared to soft labeling, which involves distillation onto the class and mask logits in our specific case. In previous approaches, the differences in actual performance might not be as apparent due to additional manipulations requiring rules and thresholds for pseudo-label generation. However, with our class-independent transformation (CIT) process, the distillation procedure is conducted more directly and is generally applicable to existing semantic segmentation methods.

\begin{table}[h!]
\centering
\resizebox{0.4\textwidth}{!}{
\begin{tabular}{ccc}
\toprule
Method(mIoU)    & DeepLabV3 & SegViTv2 \\
\midrule
pseudo label  &  39.85              &  50.80      \\
soft  label & \textbf{43.02} (+3.17) &  \textbf{53.35}(+2.55) \\
\bottomrule
\end{tabular}
}
\caption{Comparison of the performance contrast between pseudo labeling and soft labeling. The experiment was conducted under identical conditions using the ADE20K dataset following the 100-50 (2 tasks) configuration.
 }
\label{tab:pseudo}
         \vspace{-2em} %
\end{table}

\subsubsection{Accumulative pipeline vs Iterative pipeline }
We proceeded to evaluate the efficacy of the accumulative training pipeline, a capability uniquely facilitated by our CIT process, in comparison with prior methodologies that lack class-independent predictions and can only train the pipeline iteratively. We utilize the ADE20K dataset and configure our experiments using the 100-10 setup, which comprises 6 distinct tasks. The data in Tab.\ref{tab:accu} clearly show that the accumulative pipeline has superior performance. This suggests that repeated knowledge distillation on already learned categories can enhance error accumulation and degrades performance.

\begin{table}[h!]
\centering
\resizebox{0.4\textwidth}{!}{
\begin{tabular}{lcc}
\toprule
Method(mIoU)   & DeepLabV3 & SegViTv2 \\
\midrule
iterative  &   38.61          & 49.32   \\
accumulative & \textbf{41.13} (+2.52)            & \textbf{51.46} (+2.14) \\
\bottomrule
\end{tabular}
}
\caption{Comparison of the performance disparity between the accumulative pipeline and the iterative pipeline. The experiment was conducted using the ADE20K dataset under the 100-10 (6 tasks) configuration.
}
\label{tab:accu}
         \vspace{-1em} %

\end{table}

%% file: tables/exp_ade20k.tex
\begin{table*}[ht]
\centering
\label{tab:comparison on ade20k}

\caption{Comparison of class-incremental semantic segmentation results on ADE20k under the \textit{overlapped} setting. The performance of the Oracle reflects the semantic segmentation performance in fully supervised settings.}
\resizebox{0.92\textwidth}{!}{%
\begin{tabular}{lccccccccccccccc}
\toprule
& \multicolumn{3}{c}{100-50(2 tasks)} && \multicolumn{3}{c}{50-50(3 tasks)} && \multicolumn{3}{c}{100-10(6 tasks)} &&
\multicolumn{3}{c}{100-5(11 tasks)}\\
\cmidrule{2-4} \cmidrule{6-8} \cmidrule{10-12} \cmidrule{14-16}
Method & 1-100 & 101-150 & all && 1-50 & 51-150 & all && 1-100 & 101-150 & all && 1-100 & 101-150 & all \\
\midrule
Oracle DeeplabV3  %
& 47.17 & 35.79 & 43.40 &
& 53.15 & 38.53 & 43.40 &
& 47.17 & 35.79 & 43.40 &
& 47.17 & 35.79 & 43.40 \\  
Oracle Mask2former  %
& 56.32 & 47.71 & 53.45 & 
& 61.76 & 49.29 & 53.45 &
& 56.32 & 47.71 & 53.45 &
& 56.32 & 47.71 & 53.45
\\
Oracle SegVitv2 %
&56.13 &48.74 &53.67 &  
&62.95 &49.03 & 53.67& 
&56.13 &48.74 &53.67 & 
&56.13 &48.74 &53.67   \\

\bottomrule
\rowcolor{mygray} \multicolumn{16}{c}{Base method: DeepLabV3; Backbone: Resnet50} \\
\toprule
MiB~\cite{cermelli2020modeling} 
& 40.52 & 17.17 & 32.79 &
& 45.57 & 21.01 & 29.31 &
& 38.21 & 11.12 & 29.24 &
& 36.01 & 5.66 & 25.96\\
SDR~\cite{michieli2021continual}  
& 37.40 & 24.80 & 33.20 &
& 40.90 & 23.80 & 29.50 &
& 28.90 & 7.40 & 21.70 &
& - & - & -  \\
PLOP~\cite{douillard2021plop} 
& 41.66 & 15.42 & 32.97 &
& 47.75 & 21.60 & 30.43 &
& 39.42 & 13.63 & 30.88 &
& 39.11 & 7.81 & 28.75 \\
REMIND~\cite{reminder}  
& 41.55 & 19.16 & 34.14 &
& 47.11 & 20.35 & 29.39 &
& 38.96 & 21.28 & 33.11 &
& - & - & -  \\
RC~\cite{zhang2022representation} 
& 42.30 & 18.80 & 34.50 &
& 48.30 & 25.00 & 32.50 &
& 39.30 & 17.60 & 38.90 &
& 38.50 & 11.50 & 29.60 \\
SPPA~\cite{lin2022continual} 
& 42.90 & 19.90 & 35.20 &
& 49.80 & 23.90 & 32.50 &
& 41.00 & 12.50 & 31.50 &
& - & - & - \\
RBC~\cite{zhao2022rbc} 
& 42.90 & 21.49 & 35.81 &
& 49.59 & 26.32 & 34.18 &
& 39.01 & 21.67 & 33.27 &
& - & - & - \\
\textbf{CIT-DeepLabV3 (Ours)}  
& \textbf{46.32}& \textbf{36.42}& \textbf{43.02}&  %
&\textbf{53.21}& \textbf{37.20}&\textbf{42.54}&   %
&\textbf{46.12}&\textbf{31.14}&\textbf{41.13}&  %
&\textbf{45.94}&\textbf{29.40}&\textbf{40.43}\\ %
\bottomrule
\rowcolor{mygray} \multicolumn{16}{c}{Base method: Mask2former; Backbone: Swin-Base}\\
\toprule
\textbf{CIT-M2Former (Ours)}
& \textbf{56.87} & \textbf{43.53} & \textbf{52.43} & %
& \textbf{61.87} & \textbf{46.04}  & \textbf{51.32} & %
& \textbf{56.88} & \textbf{38.18} & \textbf{50.65} &
& \textbf{55.54} & \textbf{32.18} & \textbf{47.72}
 \\
\bottomrule
\rowcolor{mygray} \multicolumn{16}{c}{Base method: DeepLabV3, Segmentor, SegViTV2; Backbone: ViT-Base} \\
\toprule

MiB~\cite{cermelli2020modeling}
& 46.40 & 34.95 & 42.58 &
& 52.21 & 35.56 & 41.11 &
& 42.95 & 30.80 & 38.90 &
& 40.21 & 26.59 & 35.67 \\
Incrementer~\cite{shang2023incrementer} 
& 49.42 & 35.62 & 44.82 &
& 56.15 & 37.81 & 43.92 &
& 48.47 & 34.62 & 43.85 &
& 46.93 & 31.31 & 41.72 \\

SegViTv2 ~\cite{segvitv2}
&53.64&40.00&49.09&
&59.81&42.45&48.24&
&53.77&35.54&47.70&
&53.37&31.26&46.00 \\

\textbf{CIT-SegViTv2 (Ours)}
& \textbf{56.40}& \textbf{47.26} & \textbf{53.35} &   %
& \textbf{62.35} & \textbf{49.17} & \textbf{53.57} &  %
& \textbf{55.53} & \textbf{43.32} & \textbf{51.46}& 
& \textbf{54.60} & \textbf{37.78} & \textbf{47.66}
\\

\bottomrule

\end{tabular}%
}
\label{tab:CIT}

\vspace{-1em} %

\end{table*}

%% file: tables/exp_voc12.tex
\begin{table*}[htbp!]
    \centering
    \begin{minipage}[b]{0.66\linewidth}
	\centering
	\resizebox{1.0\textwidth}{!}{
		\begin{tabular}{lccccccccccc}
			\toprule
			\multirow{2}{*}{Method}&\multicolumn{3}{c} {19-1 (2 tasks)}&&\multicolumn{3}{c}{15-5 (2 tasks)} &&\multicolumn{3}{c}{15-1 (6 tasks)}\\ 
			\cline{2-4}\cline{6-8}\cline{10-12} \addlinespace[5pt]
			&0-19&20&\emph{all}&
			&0-15&16-20&\emph{all}&
			&0-15&16-20&\emph{all} \\
			\midrule
   Oracle ResNet50 
   & 80.71 & 85.29 & 80.94 & 
   & 81.64 & 78.83 & 80.94 &
   & 81.64 & 78.83 & 80.94
   \\
       Oracle ViT-Base	  
      &86.78&88.15&86.77&
			&86.96&86.52&86.85&
			&86.96&86.51&86.85\\
   \bottomrule
   \rowcolor{mygray} \multicolumn{12}{c}{Base method: DeepLabV3; Backbone: ResNet50} \\
   \toprule
			PI~\cite{zenke2017continual} &7.50&14.00&7.80&
			&1.60&33.30&9.50&
			&0.00&1.80&0.50\\
			EWC~\cite{kirkpatrick2017overcoming} &26.90&14.00&26.30&
			&24.30&35.50&27.10&
			&0.30&4.30&1.30\\
			RW~\cite{chaudhry2018riemannian} &23.30&14.20&22.90&
			&16.60&34.90&21.20&
			&0.00&5.20&1.30\\
			
            LwF~\cite{li2017learning} &51.20&8.50&49.10&
			&58.90&36.60&53.30&
			&1.00&3.90&1.80\\
		LwF-MC~\cite{rebuffi2017icarl} &64.40&13.30&61.90&
			&58.10&35.00&52.30&
			&6.4&8.40&6.90\\
			ILT~\cite{michieli2019incremental} &67.75&10.88&65.05&
			&67.08&39.23&60.45&
			&8.75&7.99&8.56\\
            RECALL~\cite{maracani2021recall} 
            & 67.90 & 53.50 & 68.40 &
            & 66.60 & 50.90 & 64.00 &
            & 65.70 & 47.80 & 62.70 \\
            MiB~\cite{cermelli2020modeling} &71.43&23.59&69.15&
			&76.37&49.97&70.08&
			&34.22&13.50&29.29\\
		SDR~\cite{michieli2021continual}
             & 69.10 & 32.60 & 67.40 &
             & 75.40 & 52.60 & 69.90 & 
             & 44.70 & 21.80 & 39.20 \\
             PLOP~\cite{douillard2021plop} &75.35&37.35&73.54&
			&75.73&51.71&70.09&
			&65.12&21.11&54.64\\
            REMIND~\cite{reminder} 
                & 76.48 & 32.34 & 74.38 &
                & 76.11 & 50.74 & 70.07 &
                & 68.30 & 27.23 & 58.52 \\
            RC ~\cite{zhang2022representation}
              & - & - & - &
              & 78.80  & 52.00  & 72.40  &
              & 70.60  & 23.70  & 59.40 \\
            SPPA~\cite{lin2022continual}  
              & 76.50  & 36.20  & 74.60  &
              & 78.10  & 52.90  & 72.10  &
              & 66.20  & 23.30  & 56.00  \\
            RBC~\cite{zhao2022rbc}
              & 77.26  & 55.60  & 76.23  &
              & 76.59  & 52.78  & 70.92  &
              & 69.54  & 38.44  & 62.14 \\

    \textbf{CIT-DeepLabV3 (Ours)}
    & \textbf{80.92} & \textbf{79.30} & \textbf{80.52} &
    & \textbf{80.75} & \textbf{78.35} & \textbf{80.16} &
    & \textbf{81.35} & \textbf{77.32} & \textbf{80.34}
    \\
   \bottomrule
   \rowcolor{mygray} \multicolumn{12}{c}{Base method: DeepLabV3, Segmentor, SegViTv2; Backbone: ViT-Base} \\
   \toprule
            
            MiB 
              & 79.91 & 47.70 & 79.10 &
              & 78.62 & 63.10 & 75.62 &
              & 72.55 & 23.14 & 61.73 \\
            RBC 
              & 80.24 & 38.79 & 78.99 &
              & 78.86 & 62.01 & 75.53 &
              & 75.90 & 40.15 & 68.24 \\
            Incrementer ~\cite{shang2023incrementer} 
              & 82.54 & 60.95 & 82.14 &
              & 82.53 & 69.25 & 79.93 &
              & 79.60 & 59.56 & 75.55 \\
		\textbf{CIT-SegViTv2 (Ours)} &\textbf{86.22}&\textbf{84.47}&\textbf{86.38}&
			&\textbf{86.61}&\textbf{85.27}&\textbf{86.38}&
			&\textbf{86.58}&\textbf{84.12}&\textbf{85.67}\\

		\bottomrule	
			
		\end{tabular}
	}

\caption{CSS results under the \emph{Overlapped} settings of VOC-19-1, VOC-15-5 and VOC-15-1 benchmarks. }

	\label{tab:1-2}
    \end{minipage}
    \hfill
    \begin{minipage}[b]{0.33\linewidth}
        \centering
        \includegraphics[width=\linewidth]{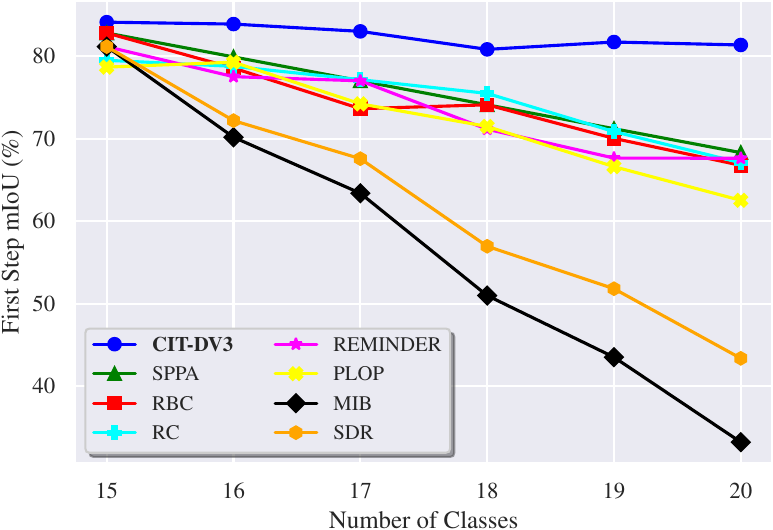}
        \captionof{figure}{DeepLabV3 19-1 (5 tasks)}
        \label{fig:image1_label_voc}
        
        \vspace{3pt} %
        \includegraphics[width=\linewidth]{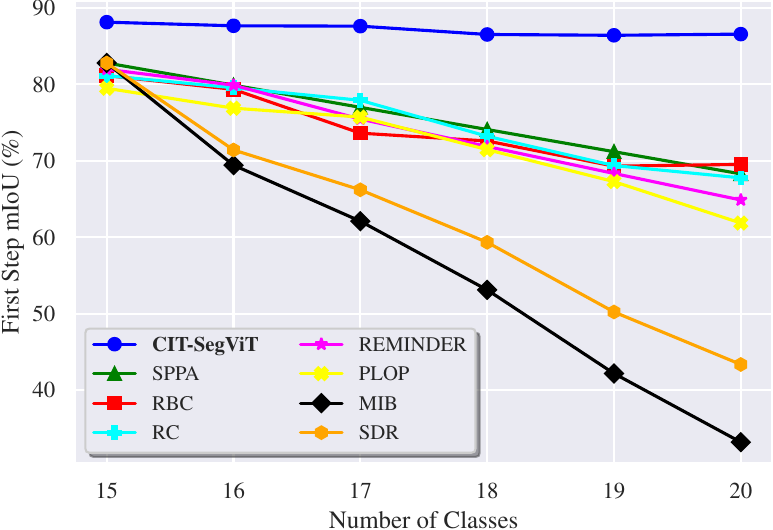}
        \captionof{figure}{SegViTv2 19-1 (5 tasks)}
        \label{fig:image2_label_voc}
    \end{minipage}
             \vspace{-1.8em} %

\end{table*}

%% file: sections/5_conclusion.tex
\vspace{-0.8em}
\section{Conclusion}
In this paper, we explore the significant challenges encountered by current CSS methods, particularly the issue posed by class-dependent predictions inherited from conventional semantic segmentation tasks. While previous CSS methods have focused on balancing interdependent probabilities between old and new tasks without forgetting, we introduce a class-independent transformation (CIT) process to circumvent this difficulty.  The CIT process, which maintains the performance of semantic segmentation, avoids complexity and instead introduces direct distillation and an accumulative training pipeline, proving to be a simpler and more effective approach for CSS tasks. Experimental results demonstrate that the class-independent approach significantly outperforms traditional methods in terms of performance. Notably, our framework enables negligible forgetting (by up to 1\%) on the base task across eight CSS settings, highlighting the effectiveness of the proposed accumulative distillation. %